\documentclass[sigconf]{acmart}

\usepackage{booktabs} 
\usepackage{amsmath}
\usepackage{amsfonts}
\usepackage[lined, algoruled, algosection, linesnumbered, resetcount]{algorithm2e}
\usepackage{paralist}
\usepackage[noend]{algpseudocode}
\usepackage{graphbox,graphicx}
\usepackage{subfig}
\usepackage{stfloats}
\usepackage{hyperref}
\usepackage{cleveref}

\copyrightyear{2019} 
\acmYear{2019} 
\setcopyright{acmcopyright}
\acmConference[GECCO '19]{Genetic and Evolutionary Computation Conference}{July 13--17, 2019}{Prague, Czech Republic}
\acmBooktitle{Genetic and Evolutionary Computation Conference (GECCO '19), July 13--17, 2019, Prague, Czech Republic}
\acmPrice{15.00}
\acmDOI{10.1145/3321707.3321829}
\acmISBN{978-1-4503-6111-8/19/07}

\begin{document}
\title[Surrogate Models for Enhancing the Efficiency of Neuroevolution in RL]{Surrogate Models for Enhancing the Efficiency of Neuroevolution in Reinforcement Learning}

\author{J\"org Stork}
\orcid{1234-5678-9012}
\affiliation{%
  \institution{TH K\"oln}
  \streetaddress{Steinmüllerallee 1}
  \city{Gummersbach} 
  \country{Germany}
  \postcode{51643}
}
\email{joerg.stork@th-koeln.de}

\author{Martin Zaefferer}
\orcid{1234-5678-9012}
\affiliation{%
  \institution{TH K\"oln}
  \streetaddress{Steinmüllerallee 1}
  \city{Gummersbach} 
  \country{Germany}
  \postcode{51643}
}
\email{martin.zaefferer@th-koeln.de}

\author{Thomas Bartz-Beielstein}
\orcid{1234-5678-9012}
\affiliation{%
  \institution{TH K\"oln}
  \streetaddress{Steinmüllerallee 1}
  \city{Gummersbach} 
  \country{Germany}
  \postcode{51643}
}
\email{thomas.bartz-beielstein@th-koeln.de}

\author{A. E. Eiben}
\orcid{1234-5678-9012}
\affiliation{%
  \institution{Vrije Universiteit Amsterdam}
  \streetaddress{De Boelelaan 1105}
  \city{Amsterdam} 
  \country{Netherlands}
  \postcode{1081 HV}
}
\email{a.e.eiben@vu.nl}



\renewcommand{\shortauthors}{J. Stork et al.}

\begin{abstract}
In the last years, reinforcement learning received a lot of attention.
One method to solve reinforcement learning tasks is Neuroevolution, where neural networks are optimized by evolutionary algorithms.
A disadvantage of Neuroevolution is that it can require numerous function evaluations, while not fully utilizing the available information from each fitness evaluation.
This is especially problematic when fitness evaluations become expensive.
To reduce the cost of fitness evaluations, surrogate models can be employed to partially replace the fitness function.
The difficulty of surrogate modeling for Neuroevolution is the complex search space and how to compare different networks.
To that end, recent studies showed that a kernel based approach, particular with phenotypic distance measures, works well.
These kernels compare different networks via their behavior (phenotype) rather than their topology or encoding (genotype).
In this work, we discuss the use of surrogate model-based Neuroevolution (SMB-NE) using a phenotypic distance for reinforcement learning.
In detail, we investigate 
a) the potential of SMB-NE with respect to evaluation efficiency and
b) how to select adequate input sets for the phenotypic distance measure in a reinforcement learning problem.
The results indicate that we are able to considerably increase the evaluation efficiency using dynamic input sets.
\end{abstract}

%
%



\maketitle

\section{Introduction}
Neuroevolution (NE) is a technique concerned with the construction of Artificial Neural Networks (ANNs) via evolutionary optimization algorithms.
One important application of NE is Reinforcement Learning (RL), where it is a considerable challenge
 to evolve competitive ANNs with evolutionary methods.
The mapping from the genotypical representation, its phenotypic behavior, and finally
to the fitness measurement (i.e., its ability to solve a learning task) can become extremely complex.

Evolutionary algorithms will need to spend a significant amount of fitness function evaluations
to find well-performing networks. This may become an issue if fitness evaluations
are expensive and dominate the overall time or resource consumption of the optimization process.
Surrogate Model-Based Optimization (SMBO) is one way to deal with this issue \cite{Jin2011}.
Here, data-driven models partially replace the expensive fitness function.
Except for few recent studies~\cite{gaier2018data,Stor17c,Stor18a,Stor18c}, SMBO has found no application
in the context of NE.

Following these recent developments, we intend to design surrogate models that 
allow to learn a cheap yet accurate representation of the genotype-phenotype-fitness mapping.
In that context, we also focus on kernel-based Kriging models.
The approach of kernel-based modeling with Kriging for complex, combinatorial structures
is discussed in more detail by Zaefferer~\cite{Zaefferer2018c}. 

For graphs, such as ANNs, this can become a difficult task.
Specific graphs, such as trees, may allow computing kernels based on measures like the tree edit distance~\cite{Pawlik2015}.
Such distances on the genotype can be plugged into the kernel function (i.e., replacing the Euclidean distance) and used to model the genotype-fitness mapping~\cite{Zaef18b}.
However, the same is not as simple for graphs like neural networks, as the computation of edit distances is NP-hard in the general case.
At best, approximate distances can be used, such as the compatibility distance employed by Gaier et al.~\cite{gaier2018data}.
Stork et al. \cite{Stor17c} discuss the use of surrogates of fixed ANN topologies in control tasks using genotypic distances.

As an alternative to genotypic distances, it is often possible to compute the distance on some form of behavior or phenotype.
This idea was first discussed by Hildebrandt and Branke~\cite{Hildebrandt2014} in the context of genetic programming for dynamic job shop scheduling problems. 
It was later also tested for symbolic regression by Zaefferer et al.~\cite{Zaef18b}.
The key idea of this approach is that complex structures can be compared by observing their output (phenotype), rather than their structure (genotype).
In terms of NE, different distance measures were recently tested for classification problems by Stork et al.~\cite{Stork2018a,Stor18c}. 
They came to the conclusion that phenotypic distances for ANNs are promising if the correct input signal is chosen.

A fairly different model has been used for a RL problem in the context of NE by 
Koppejan and Whiteson~\cite{Koppejan2011}.
Their goal was not to replace the fitness function (as is usually done in SMBO). Rather, they intended to reduce the sample
cost involved in testing a controller within different instances of a specific problem class.
Instead of a purely data-driven model, they employ a model that is mostly based on the understanding of the physical system under consideration (a hovering helicopter).
In terms of the classification by Bartz-Beielstein and Zaefferer~\cite{Bartz-Beielstein2016n}, this can be seen as a customized modeling strategy. A transfer to other problem domains is not straight-forward.
Most approaches considered in our study can be seen as similarity-based strategies or mapping strategies.

In contrast to the related work, we focus on the application of Surrogate Model-Based NeuroEvolution (SMB-NE) for RL and the special needs which arise from solving such tasks, particular considering the computation of a phenotypic distance. 
In summary, we investigate the following questions:
\begin{enumerate}[Q-I]
\item How can we learn the mapping from a neural network to its performance in terms of solving RL tasks?
\item How does a model-based NE compare to model-free NE on RL problems?
\item How should phenotypic distances be configured to generate well-performing surrogate models?
\end{enumerate}

We describe the corresponding models and algorithms for NE and RL in \cref{sec:methods}.
SMB-NE in the context of RL is described in \cref{sec:smbrl}.
Our experimental setup is described in \cref{sec:experiments} and the results are discussed in \cref{sec:results}.
A final summary and an outlook on future work are given in \cref{sec:conclusion}.

\section{Methods}\label{sec:methods}
\label{sec:rlearning}
The application of model-based search for RL introduces a set of challenges. 
In general, for solving RL problems, we want to distinguish between three possible scenarios:
\begin{enumerate}[S-1]
\item \emph{New task:} We want to solve a new task, i.e., no prior experiments were conducted and no data is available. Here, no information from prior runs can be used to accelerate the current run and initial experiments have to be conducted to gather information. 
\item \emph{Same task, different instance:} Data and optimized controllers from former experiments are available and a different problem instance of the same envionment (e.g. different start parameters) has to be solved. For many of these cases, the ANN controller trained for prior runs may be reused if it is not overfitted and provides a robust solution performance. If not, the existing ANN controller can be subject to further optimization, where a fast convergence to a good solution is anticipated. 
\item \emph{Same task, different environment:} Data from former experiments is available, but a different, yet similar environment needs to be solved. For example, changes to the environment, such as a different maze (in a maze solving problem) or different physical shape of a robot or appliance, could be considered. 
In this case, a prior optimized ANN controller could provide a good starting solution. An available data model of the optimization run could still provide valid information.
\end{enumerate} 
In this work, we will focus on the first scenario (S-1), whereas for (S-2) different instances will be tested and a robust controller is part of the benchmark target. (S-3) will be part of future work. 

Our model-based approach for solving RL tasks is a combination of existing algorithms: Cartesian Genetic Programming for Neural Networks (CGP-ANN), Surrogate Model-Based Optimization (SMBO), and specific Kriging models utilizing Phenotypic Distance (PhD) kernels. We describe these algorithms in the following.

\subsection{Cartesian Genetic Programming for Neural Networks}
\begin{figure}[t]
\centering
\includegraphics[width=0.45\textwidth]{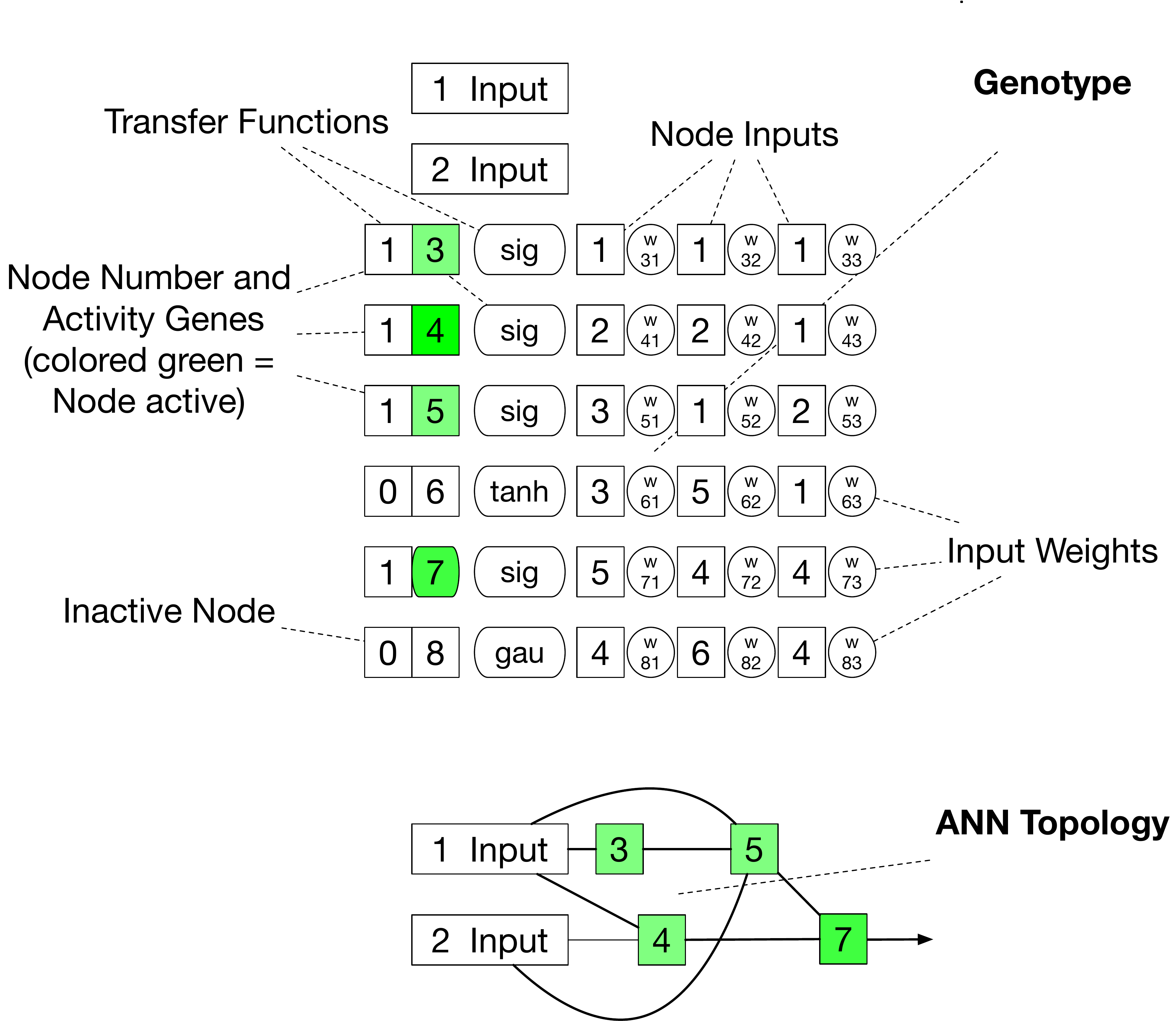}
\caption{A CGPANN genotype with two inputs, eight nodes, an arity of three and different transfer functions. Each node has a transfer function, a boolean activity gene and several inputs with adjacent weights. Green nodes are active and part of the encoded ANN. In the related topology only active nodes are included and duplicate connections are aggregated. Taken from \cite{Stor18c}.}
\label{fig:chromosome}
\end{figure}

In this work, the ANNs are encoded as in the CGP-ANN algorithm \cite{khan2010,miller2000cartesian,turner2013cartesian}.
Each individual consists of a fixed number of nodes, as visualized in \cref{fig:chromosome}.
Beside the input nodes, which represent the data inputs, each node has a single transfer function, a fixed number of inputs and associated weights based on their arity. 
Nodes are always connected to proceeding nodes and multiple connections to the same node are possible. 
Only those nodes which are directly or indirectly connected to an output are evaluated during a run, while all other remain passive and do not influence the behavior of a network. 
Thus, very small active ANN topologies and also those only using specific inputs are possible, even if the genotype has numerous nodes. 
In CGP-ANN, the networks are optimized using mutation, following the concept an Evolutionary Strategy (ES). 
A typical choice is a (1+4)-ES, whereby the elitist is always the current best individual (e.g., if during evolution an individual achieves the same fitness as the elitist, the more recent is selected). 
The C library \texttt{CGP} by A.~Turner\footnote{http://www.cgplibrary.co.uk - accessed: 2018-01-12}, extended by interfaces to R, was used to perform the experiments. 

\subsection{Kriging for modeling ANNs}\label{sec:kriging} 
In this study, we focus on an SMBO approach that employs Kriging (Gaussian process regression)~\cite{Forrester2008a}.
The main question in this context is how Kriging can model the complex dependencies between a neural network's 
topology and its fitness. 

At its core, Kriging is based on kernels such as the exponential kernel $\text{k}(x,x')=\exp(-\sum_{i=1}^{n}\theta_i (x_i-x_i')^2)$. In this example, $x \in R^n$ is a vector of real values, and $\theta_i \in R^+$ is a non-negative parameter of the kernel.
If $x$ is not a real valued vector, but rather represents a candidate ANN, we need to change the kernel
such that it compares networks rather than vectors. 
For instance, the weighted distance measure $-\sum_{i=1}^{n}\theta_i (x_i-x_i')^2$ may be replaced with some distance between ANNs.
To that end, previous work suggested evaluating distances that are based on observations of network
behavior (or phenotypes)~\cite{Hildebrandt2014,Zaef18b,Stor18c}. 
More details on the computation of phenotypic distances are given in~\cref{subsec:phd}.

One complication of using such phenotypic distances in Kriging is dimensionality.
Specifically, Kriging is often suggested for problems with less than 20 variables (e.g., see Table 3.1 in \cite{Forrester2008a}).
At the same time, the vectors of phenotypes we consider in the context of ANNs may easily grow to lengths of 100 or more elements.
Hence, the combination of Kriging and phenotypic distances may appear to be a poor choice.

We propose to tackle this in two manners, each related to two different aspects that affect problems
with high-dimensionality in Kriging.
One problem of high-dimensional data is the determination of the kernel parameters such as $\theta_i$.
These are usually determined by Maximum Likelihood Estimation (MLE), using numerical optimization algorithms~\cite{Forrester2008a}.
In MLE, the parameters are chosen to maximize the likelihood determined by the Kriging model.

Clearly, the number of parameters increases with the dimensionality of the data.
Optimizing many parameters by numerical optimization may pose a very difficult problem.
A straight-forward fix is to set all parameters $\theta_i$ to the same value, that is, to use
$\text{k}(x,x')=\exp(-\theta \sum_{i=1}^{n} (x_i-x_i')^2)$, where only a single parameter $\theta$
has to be determined by MLE. 
That is, we choose an isotropic rather than anisotropic model.

\begin{figure}[t]
    \centering
        \includegraphics[width=0.7\linewidth]{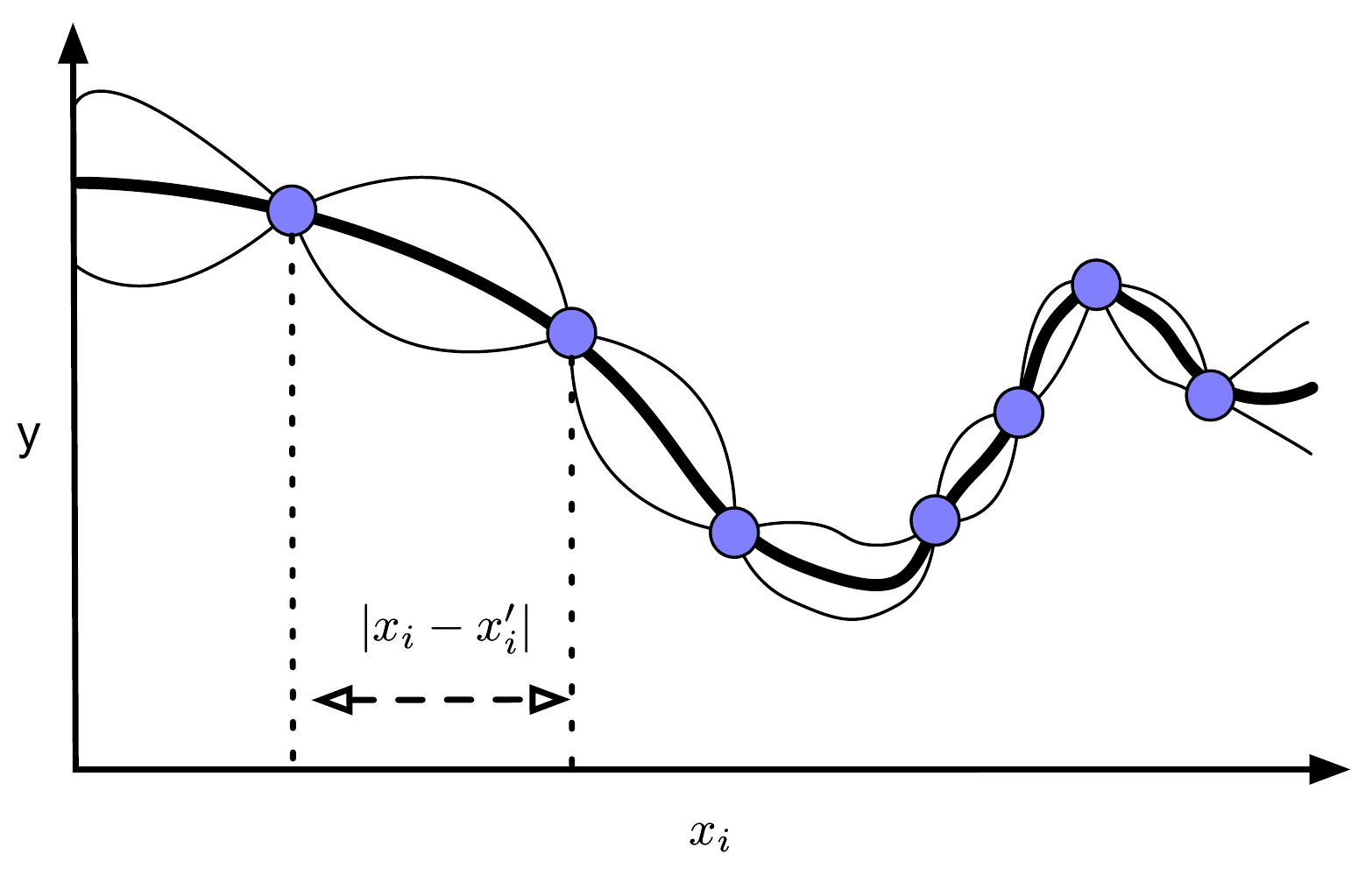}%
    \caption{Example for Kriging modeling: distances between ANNs (blue circles) in one dimension $x_i$ of a phenotypic output. The bold line is the model prediction, while the thin lines display the uncertainty of the model. }
    \label{fig:krige}
\end{figure}

A second problem is the behavior of distances in high-dimensional spaces.
Roughly speaking, when measuring Euclidean distance of different data points in a high dimensional space,
nearly all points will have the same distance~\cite{Aggarwal2001}. 
Clearly, this is undesirable for any model that is based on such distances.
Aggarwal et al.~\cite{Aggarwal2001} state that other distances are less affected by this issue, especially the Manhattan distance
(based on the $L_1$ norm).
For that reason, we finally propose to use an isotropic kernel based on the Manhattan distance to measure the distance
between phenotype vectors, i.e., 
\begin{equation}\label{eq:kphd}
\text{k}_\text{PhD}(x,x')=\exp(-\theta \sum_{i=1}^{n}|x_i-x_i'|).
\end{equation}
Figure \ref{fig:krige} illustrates an example model for a  one-dimensional case. 
Besides dimensionality, another important aspect of kernels for Kriging is their definiteness.
Usually, kernels are required to be positive semi-definite.
The kernel $\text{k}_\text{PhD}$ from \cref{eq:kphd} is definite, as it is a special case of the positive semi-definite Gaussian kernel~\cite{Forrester2008a}.

\section{Surrogate Model-based Neuroevolution for Reinforcement Learning}\label{sec:smbrl}

\subsection{Phenotypic Distance Measure for ANN Topologies}\label{subsec:phd}
Evolved ANN topologies do not have fixed structures in terms of hidden layers, weights, connections, or functions.
Measuring a distance between these complex structures is thus a challenging task.
In detail, it is difficult to measure a distance directly on these structures, due to several problems:

\begin{enumerate}[P-1]
\item \emph{Competing Conventions:} A famous problem which arises in the context of ANNs are competing conventions~\cite{schaffer1992combinations}, i.e., different genotypes can result in the same topology, as well as the same phenotype.
\item \emph{Incomparability:} Even ANNs with fixed genotypic structures (e.g., as produced by CGP), are often not directly comparable. When comparing two genotypes, elements such as certain nodes may be not aligned in the same way, despite having the same effect on the outcome. In other words, it is not always straightforward to decide which pairs of nodes should be aligned with each other, when comparing two different networks.
This could be handled by complex and computationally expensive sorting and aligning processes, but this would pose an optimization problem in itself and render the comparison computationally expensive \cite{Stor18c}. This issue becomes more problematic with increasing size of the considered networks.
\item \emph{Lack of Smoothness:} 
In some cases, small changes in the genotype can have a significant effect on the final behavior.
For instance, removing a single connection may change the fitness of the network dramatically.
That means, small distances in the search space may lead to large distances between fitness values.
Essentially, this implies that the search space is not smooth under a genotypic distance. 
This presents a severe problem to every optimization or modeling algorithm.
\item \emph{Distance Balancing:} Different types of changes in a network have different meaning and impact. 
For example, changing a weight has a different impact than changing the transfer function of a node.
It is thus difficult to provide a meaningful distance that correctly balances (or weights) such changes. 
For example, it is unclear whether two networks that only differ in a single weight are at the same distance as two
networks that only differ in a single transfer function. 
\end{enumerate}
Due to these issues, genotypic distance do not seem to be ideal for the computation of high-performing surrogate models in SMB-NE. Thus, we follow the idea of comparing the behavior of ANNs and employ a Phenotypic Distance (PhD) for modeling the dependencies of ANNs~\cite{Hildebrandt2014,Zaef18b,Stor18c}. 
In the context of ANNs, we consider the reaction (output) of an ANN to an input signal to be its behavior or phenotype.
In detail, we compute the PhD by first selecting a representative input vector for a given task. 
Then, these inputs are fed into the ANN, and we observe a vector of output values.
In terms of RL, the input vectors are typically vectors of consecutive observed environment states. 
The observed outputs are then directly compared via the kernel described at the end of \cref{sec:kriging}.

The clear advantage of the PhD is that the output length and structure are not dependent on the genotype or topology of the compared ANNs. 
The computation of the input to output mapping can be performed independently of different structures (e.g, it does not matter whether the network has 10 or 1000 active connections, different transfer functions etc.).
Moreover, the observed outputs of the PhD give a clear impression of how the networks react and explicitly account for granular changes in how ANNs differ in solving a specific task. 

The potential disadvantages of the PhD are the computation times for generating the output vectors. 
For complex, large ANNs topologies (genotype size does matter much less) they can sum up to a significant amount.
This issue is less significant if we consider the task itself to be computationally expensive, especially for expensive simulators, or even real-time experiments. 
In this study, we thus concentrate on enhancing the evaluation efficiency and not the overall computation time, which is strongly related to solving a specific task.

\subsection{Phenotypic Distances: Input Vector Selection}\label{sec:inputSelection}
The PhD is, in contrast to genotypic distances, strongly task and environment dependent. 
One critical aspect of employing phenotypic distances is thus that their design needs to be adapted to each specific application.
For our purpose of modeling ANNs in RL, we first need to define how the input of the networks are chosen.
These inputs can then be used to generate the output of the ANNs, which will be compared by the distance measure.

Choosing the input vectors involves multiple issues, such as distribution of the state data, size of the data set, and many more.
In this paper, we focus on four different types of input vectors:
\begin{itemize}
\item \textbf{Precomputed set (Pre):} As a comparison baseline, we use the set of states that are observed by optimal or near-optimal solutions. A number of state vectors from previously successful runs are stored and used as a precomputed input set. This is an artificial baseline for our outlined scenario (S-1), where the knowledge about these states is initially not available. This could be seen as a best-case scenario.

\item \textbf{Static initial set (Init):} We extract the observation vectors $s_{t=1}$ from the initial dataset $D_{t=1}$ to form the input vector $v_{t=1}$, which is not altered during the run.
The possible downside of this approach is, that the initial, randomly generated solutions typically have a poor fitness. 
Clearly, poor solutions may see quite different states than successful, near-optimal solutions.
They may cover only a small subset of all possible observable states.
As these initial input vectors are thus not representative, the resulting distance may not be useful in predicting good solutions. 
\item \textbf{Sampling set (LHS):} Furthermore, input vectors can also be determined by design of experiment methods, such as latin hypercube sampling (LHS)~\cite{mckay1979comparison}. 
The idea is to distribute the data in a space-filling manner between known bounds for the network inputs. 
In contrast to the static initial approach, these inputs cover the whole state space, they are artificial and do not represent the observed states of real runs. 
\item \textbf{Dynamic set (Dyn):} Finally, we may update $v_{t}$ at each iteration, if a new best ANN is found by the SMBO algorithm.
The observed states $s_{t}$ of this new solution will replace the worst in the input vector $v_{t}$.
This implies that the input vectors are changed dynamically over time and may approximate the baseline if the algorithm converges to the optimum.
Clearly, this also means that the employed distance measure changes in each iteration. 
Models trained in different iterations of the algorithm are hence not directly comparable.
\end{itemize}

Further considerations include the number of observations in the input vector and maximum size of overall input vector.
A large size for $v_{t}$ increases the computation time for generating the phenotypes of the ANNs and further may introduce problems with too high dimensionality (\cref{sec:kriging}). 
On the other hand, a larger input vector including the states of different runs may lead to a more representative phenotype and thus distance. 
In this context, another problem is introduced by the dynamic input vectors, as their size can change over time and is more difficult to control.

\section{Surrogate Model-based Neuroevolution for Reinforcement Learning}\label{sec:smbrl}
\begin{figure}[b]
    \centering
        \includegraphics[width=0.9\linewidth]{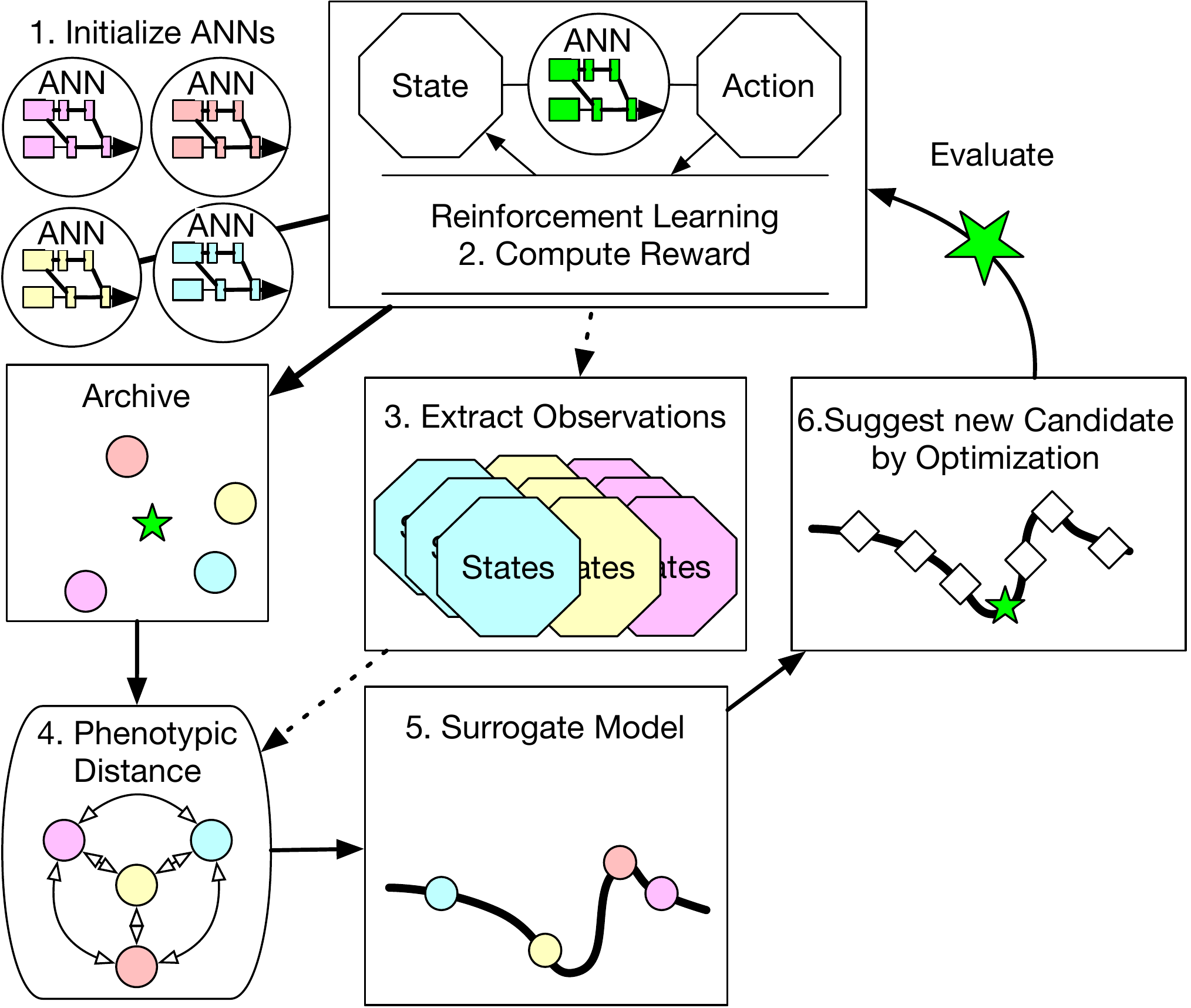}%
    \caption{SMB-NE Cycle for Reinforcement Learning}
    \label{fig:smbne}
\end{figure}

In this work we employ SMB-NE, which was first introduced by Stork et al. \cite{Stork2018a,Stor18c} and tested by 
evolving neural networks for classification problems.
The SMB-NE algorithm follows the principles of Efficient Global Optimization \cite{Jones1998}, which was introduced in the context of expensive real-world optimization problems.
In case of SMB-NE for RL, we require additional steps due to the complex ANN structures. The complete procedure is outlined in \cref{fig:smbne} and \cref{alg:smbne} .
The detailed steps of the algorithm are:
\paragraph{Initialization of model set}
The algorithm starts by sampling an initial set with $k$ candidates   $D_{t}= \{(x_{1:k},y_{1:k})\}$.
The initial set is generated completely at random, so the included genotypes can differ strongly in their size, given by the number of genomes, connections, functions and weights. The active ANN topologies are then compiled with CGP to be evaluated with the underlying RL task.
\paragraph{Evaluation with RL task}
Each ANN is evaluated over a number of time steps, which are defined by the RL task. 
In each time step, the ANN computes actions based on the currently observed state inputs.
Each action is then given a reward. 
Some actions lead to a negative reward, such as a robot bumping in a wall, or expending some resources. 
Positive rewards are given for accomplishing a certain goal. 
The fitness of a so-called episode is typically the sum of all rewards over the executed time steps. 
The fitness information is thus limited, as it includes only the final reward of an episode and does not reveal which single action was beneficial or not.
\paragraph{Extraction of observed states}
If the input vector $v_{t}$ is not given upfront or precomputed by DOE methods, it needs to be extracted from the RL experiment.
For each experiment, all observed system states are stored in a vector $s_t$. 
The set of state observation vectors is sorted according to the determined fitness values for each experiment. 
From this set, the best $num_{s_t}$ observation vectors (according to the fitness of the respective ANN) are selected. 
They are combined in the order of their fitness to form a single input vector $v_{t}$ with length $len_{v_t}=num_{s_t} * len_{s_t}$.
If the observation vector length is beyond a certain size, a subset of each $s_t$ is selected before combining them to $v_{t}$.
This intends to keep the number of elements in the vector from becoming too large.
\paragraph{Kriging model construction}
The Kriging model is constructed as described in \cref{sec:kriging}.
We utilize the R-package \texttt{CEGO}~\cite{CEGOv2.2.0} to train the Kriging model. 
At the start of the process, the model is trained with the initial set of solutions $D_{t=1}$.
The state input vector $v_{t}$ is used to compute the phenotype of each ANN, which is required to calculate the PhD.
In the later modeling steps, a subset $M_t$ is selected from $D_t$ to build the model.
This subset selection intends to avoid issues with growing data sizes, which may render
the Kriging model too time-consuming to compute. 
This set $M_t$ contains $num_{m}$ of all archived solutions. 
It is typically set to $num_{m} > 100$, so for runs with fewer than 100 evaluations it has no effect.
$M_t$ is formed by combining a number (typically $\frac{1}{5}*num_m$) of the best found solutions with the rest being sampled at random from the archive (without replacement, thus duplicates are not possible). 
This process further influences the balance between exploration and exploitation, as in each iteration a different set of ANNs is considered for the model construction. 
\paragraph{Surrogate Optimization}
The sequential optimization steps are conducted by optimizing the Expected Improvement (EI) of the surrogate model to suggest new promising ANNs.
The EI criterion delivers a balance between the predicted fitness and the uncertainty of a solution, thus also leading to a balance between exploration and exploitation in den model-based search~\cite{Jones1998}.
For the model optimization, we utilize the same (1+4)-ES of CGP-ANN to generate new candidates. 
To predict the fitness of new candidates their PhD needs to be computed, which requires their ANN outputs, based on the selected input vector $v_{t}$.
The identified candidate with highest EI on the surrogate model is again evaluated with an RL run and added to the archive $D_t$.
\paragraph{Dynamic state vector update (optional)}
If the \emph{dynamic} strategy for the input vector $v_{t}$ is chosen, it is updated if the new candidate solution has a better fitness than the best known solution.
During this update, the observed states $s_{t}$ of the related RL run replace the ones of the worst candidate solutions in the input vector $v_{t}$. 
\paragraph{Next iteration}
If the stopping criterion is not met, the next iteration is started.

\begin{algorithm}
 \caption{Surrogate Model-based Neuroevolution for Reinforcement Learning}
 \label{alg:smbne}  
 \Begin{
 $t=1$ \\
 \textbf{initialize} $k$ CGP genotypes ($x_i$) at random  \\
 \textbf{evaluate} their fitness with the objective function to get initial solutions 
 $D_{t}= \{(x_{1:k},y_{1:k})\}$ \\
 \textbf{extract} state vectors to create PhD input vector $v_{t}$ \\ 
 \textbf{build} Kriging surrogate model $m_t$ with $D_{t}$ using input vector $v_{t}$ to compute ANN phenotypes for PhD  ;   \\
 \While{\textbf{not} termination-condition}{
\If{$t > 1$}{\textbf{rebuild} surrogate model $m_t$ with selected subset $M_t \subseteq D_t$}
\textbf{optimize} EI estimated by $s_t$ with evolution strategy to discover promising $x_{t+1}$  \\
\textbf{evaluate} network $x_{t+1}$ with the objective function \\
\If{$y_{t+1} <  y_{t}$}{
 \textbf{update} input vector $v_{t+1}$ with states of successful run (dynamic input vector)  \\ 
}
\textbf{update} archive $D_{t+1}= \{D_{t},(x_{t+1},y_{t+1})\}$  \\
 $t=t+1$
 }
}
\end{algorithm}

\section{Experiments} \label{sec:experiments}
\begin{figure}[b]
    \centering
        \includegraphics[width=0.5\linewidth]{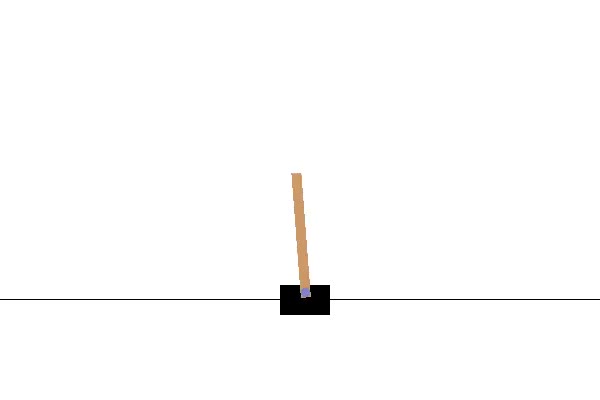}%
        \includegraphics[width=0.5\linewidth]{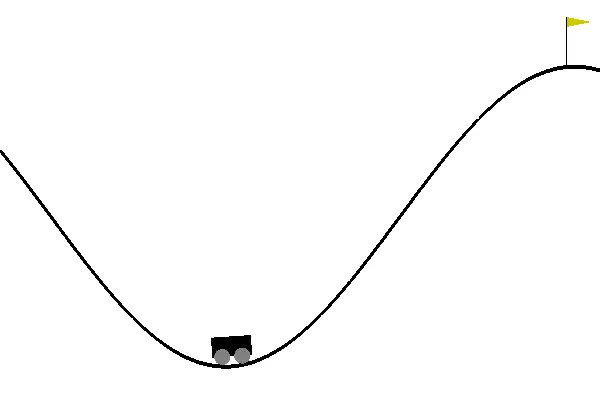}%
    \caption{OpenAI Gym CartPole-v1 and MountainCar-v0}
    \label{fig:gym}
\end{figure}

For our experiments, we chose two problems from the OpenAI Gym toolkit as a benchmark, because they are well-known in the community. Specifically, we chose the classic RL problems \emph{CartPole-v1} and \emph{MountainCar-v0}, displayed in \cref{fig:gym}..
They are implemented in python and the \texttt{reticulate} package in R was used to create an interface between SMB-NE and openAI Gym.

\subsection{OpenAi Gym Benchmarks}

The \emph{CartPole-v1} environment is a classic cart-pole balancing problem, where a pole is placed with an un-actuated joint to a cart moving on a frictionless track. 
It has four observations per state, the cart position, cart velocity, pole angle, and the pole velocity at its tip. Based on these observations, two discrete actions can be chosen by a controller, either pushing the car to the left or to the right. 
The goal is to keep the pole balanced and the cart near to the center of the track.
Each episode of the environment is evaluated over 200 time steps and terminated if the pole or cart moves out of pre-defined ranges. 
For each time step reward of 1 is given and the environment is considered solved if an average reward of 195 is achieved per episode over 100 trials. 

In the \emph{MountainCar-v1} environment, a car situated between two hills on a one-dimensional track has to be driven up a mountain, whereby
the acceleration of the car is not strong enough to drive up directly. Thus, a swinging forward and backward behavior is needed to succeed. 
The observation space consists of only two variables, the current position and velocity. The action space has three discrete options: drive left, do nothing and drive right. 
Again, an environment episode is run over 200 steps, but terminated if the goal is reached. 
For each step, a negative reward -1 is given and the environment is considered solved if a reward larger than -110 is achieved over 100 trials.

While for CartPole the fitness function is set to direct negative reward (as we utilize minimization during optimization), the MountainCar fitness function is altered for the optimization to a mixture of achieved maximum height (maxHeight) and reward by 
\begin{equation}
fitness: y(x)=- (maxHeight_{episode} + \frac{Reward_{episode}}{100})
\end{equation} 
This modification was chosen to compute a more granular fitness. 
Without this, most initial solutions get the worst reward. 
An initial data-set where nearly all solutions have the same poor fitness would be detrimental for training a surrogate model.
The stopping criterion remains unchanged, based only on the reward. 

\subsection{Parameter Tuning and Setup}
Due to the considerable runtimes, we were not able to perform exhaustive tuning of the parameter space for CGP-ANN and SMB-NE, but conducted
some preliminary tests to acquire information about the algorithm parameter space and the significance of specific variables.
For CGP-ANN, two mutation operators (\emph{single active mutation} and \emph{random mutation}) as well as different mutation strengths were tested.
The preliminary tests have shown, that CGP-ANN with a \emph{single active mutation}, where in each iteration the genotype is mutated until at least a single active genome (and an arbitrary number of non-active genomes) is altered, was not able to deliver a competitive performance. 
Moreover, the choice of the mutation strength in random mutation has a significant impact on the performance.
Thus, we decided to conduct experiments with different mutation strength of 2,5 and 10 percent, which were selected based on former experiments to show the influence of this parameter and conduct realistic comparisons. 
SMB-NE includes an even larger set of parameters, such as the choice of input vectors (for the PhD), their number and dimensions, as well as optimizer and its parameters during the surrogate model optimization. 
Most of these parameters were thus set by the authors experience and due to the small set of preliminary tests.
We identified that the number of different observation vectors $num_{s}$ for creating the input vector $v_{t}$ might be an important tuning factor and thus added different variants to the experiments.
Although we expect that the chosen parameters for both algorithms do not reflect the best possible options, they should still provide valuable insights on the performance level of both algorithms.
\begin{table}[t]
\caption{Algorithm Parameter Setup for the Experiments}
\small
\begin{tabular}{llll}
 \bf Problem & \bf Weight Range& \bf Nodes & \bf Arity  \\
CPole/MCar&   {[}-1,1{]} & 200/100 & 20/10  \\
\midrule
\bf CGP-RS  & \bf  & \bf Max Episodes & \bf  \\
CPole/MCar&&3000/5000&\\
\bf CGP-ANN  & \bf Mutation rate & \bf Max Episodes & \bf  \\
CPole & 2/5/10   & $20 + 750\cdot4$ &  \\
MCar & 2/5/10   & $20 + 1250\cdot4$ &  \\
\bf SMB-NE  & \bf PhD Input Sets & \bf Max Episodes& \bf Surr Evals  \\
CPole & Pre, Init,  LHS   &20 + 3000 & 1000 per iter\\
CPole & Dyn: $num_{s}=$ 2/5/10   &20 + 3000 & 1000 per iter\\
MCar & Pre, Init,  LHS  &20 + 5000 & 1000 per iter\\
MCar & Dyn: $num_{s}=$ 2/5/10  &20 + 5000 & 1000 per iter\\
\emph{MCar} & \emph{Dyn**}, $num_{s}=$ 5   &\emph{20 + 5000} & \emph{4000 per iter}\\
\end{tabular}
\label{tab:setup}
\end{table}

\Cref{tab:setup} shows the parameter setup for the benchmarks.  
CGP-ANN genotypes for CartPole/MountainCar are set to an arity of 20 or 10, with 200 or 100 nodes, resulting in up to 
4000 or 1000 connections between nodes.
We perform all test with a large set of activation functions: tanh, softsign, step, sigmoid, and gauss. 
Both, the maximum size of the genome, and the function were set by the authors experience. 
This displays a typical scenario in NE, where we do not know upfront which size for the genotypes is best. 

All inputs are, if possible, normalized to the [-1,+1] range and the connection weight range was also set to [-1,+1].
The setup considers a maximum runtime of 3000 or 5000 episodes, whereby the run is stopped as soon as the stopping criterion (environment solved) is met. 
The size of the initial data set $D_{t=1}$ is set to 20 for all algorithms. 
CGP-ANN starts the normal evolution with the best found solution of the initial set and computes four candidates per iteration. 
SMB-NE selects a single new candidate per iteration and utilizes 1000 search steps for the model optimization. 
All SMB-NE setups use the same mutation rate during the model search (5\%).
The tested setups include all variants introduced in \cref{sec:inputSelection} (\emph{Pre, LHS, Init} and \emph{Dyn}).
For the dynamical approach, different numbers of observation vectors $num_{s}$ to create the input vector $v_t$ are tested. 
The input vectors generated by LHS are based on the (theoretical) bounds of the state observations for each environment and have 800 elements. 
All experiments are repeated 30 times with different random number generator seeds.
Different algorithms/configurations are tested with the same set of seeds to be comparable. 
A CGP-ANN configuration that only generates random solutions is included in the experiments as a baseline (RS).
For assessing the performance of an exhaustive model search, we test an additional variant (Dyn**) of SMB-NE with the dynamic set for MountainCar-v0, where we set the size of the surrogate model evaluations to 4000.  Due to the computational effort, this variant is only repeated 20 times. 
The statistical significance of the observed differences are evaluated using the Kruskal-Wallis rank sum test~\cite{Kruskal1952} and a posthoc test for multiple pairwise comparisons according to Conover~\cite{Conover1979}.
\section{Results and Discussion}\label{sec:results}
\begin{figure*}[t]
    \centering
        \subfloat[CartPole-v1]{%
        \includegraphics[width=0.49\linewidth]{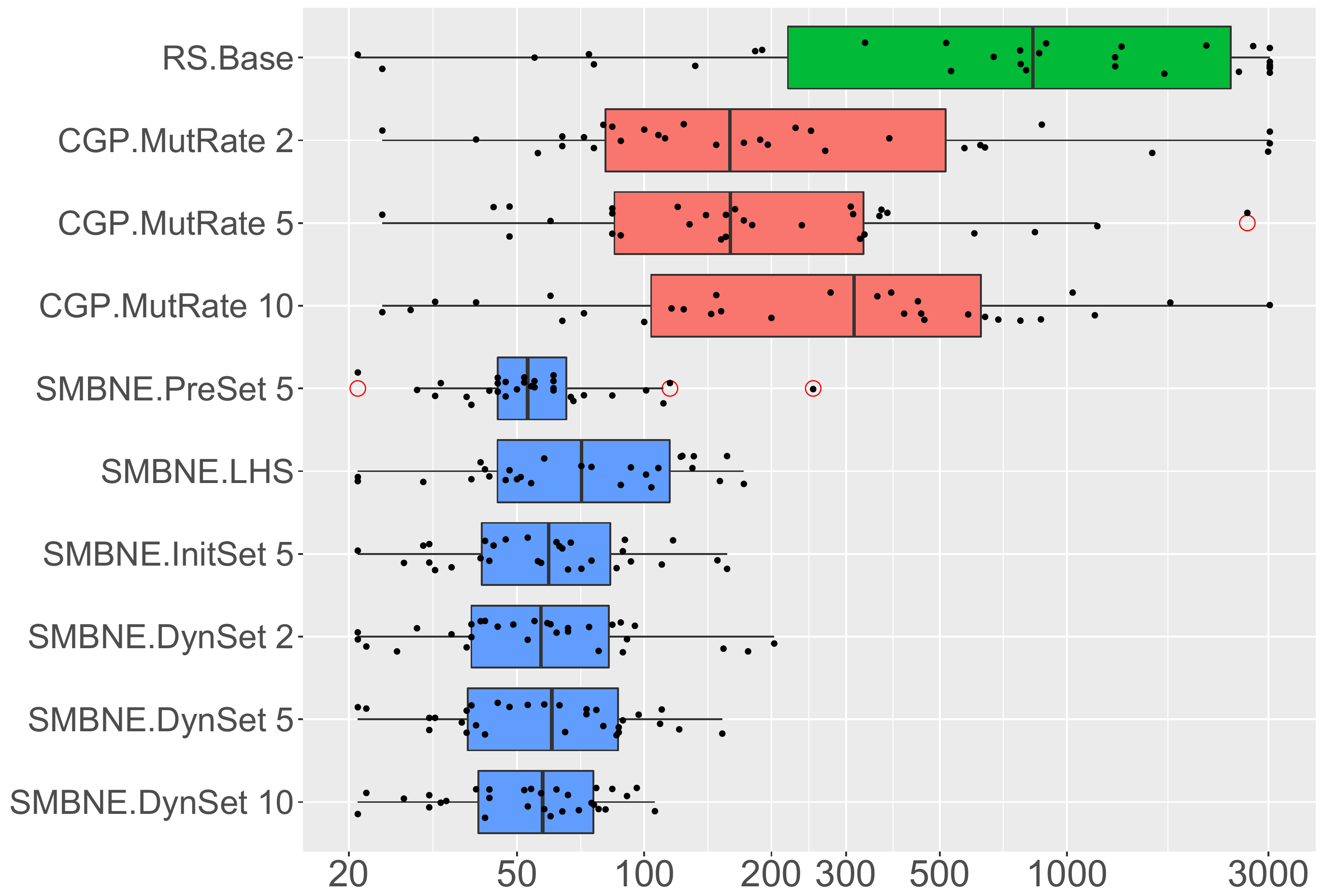}%
        \label{fig:rescartpole}%
        }%
    \hfill%
    \subfloat[MountainCar-v0]{%
        \includegraphics[width=0.49\linewidth]{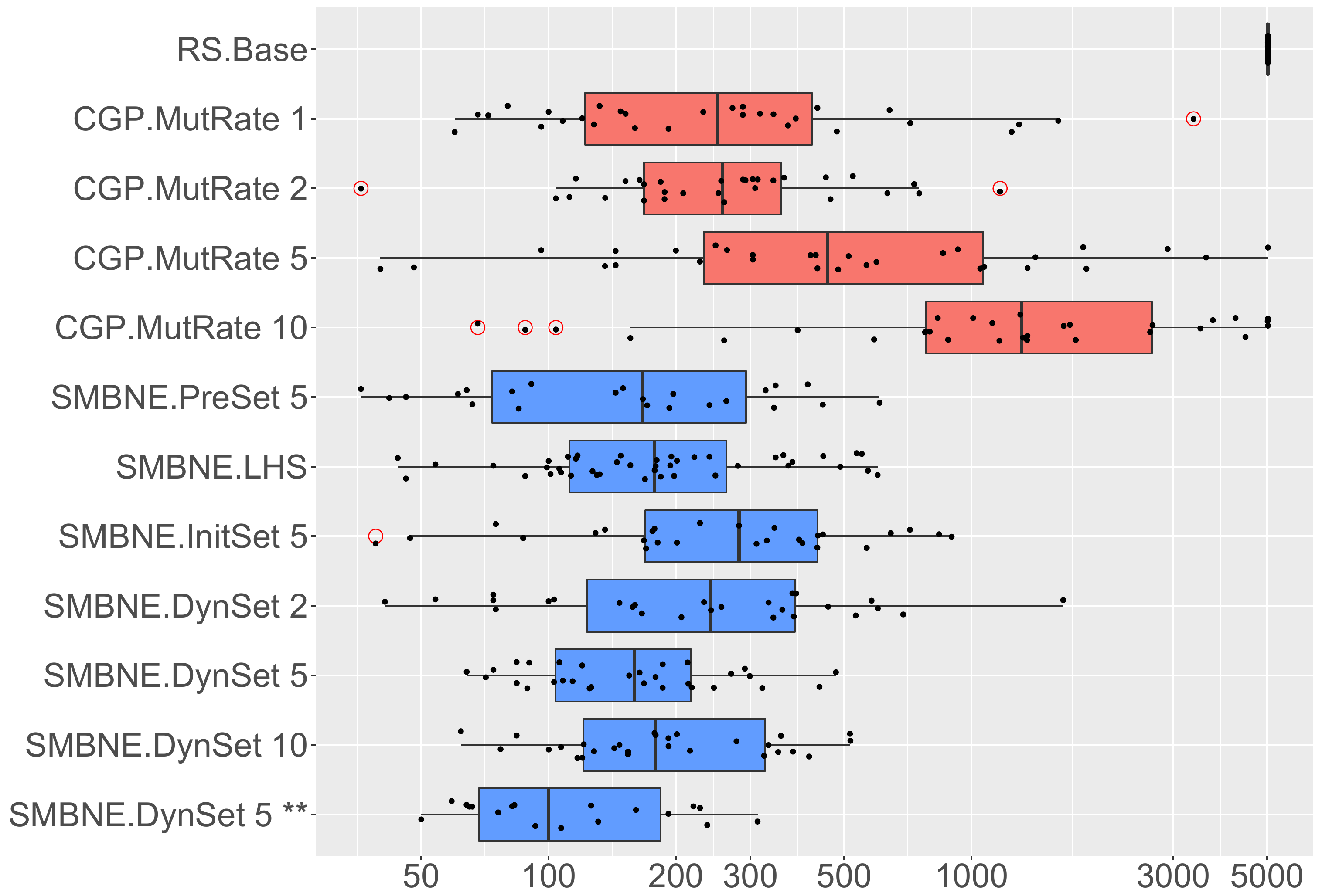}%
        \label{fig:resmountcar}%
        }%
    \caption{Experimental results. The number of required function evaluations (episodes) to solve the environments is Log10 scaled. Algorithms have different colors and specific setups are attached to the algorithm names. The numbers indicate either the utilized mutation rate in percent (CGP-ANN) or the number of utilized state observation vectors $num_{s}$ (SMB-NE). }
\end{figure*}

\begin{table}[hb]
\caption{Result tables for both environments, reported mean and standard deviation, sorted by CartPole-v1 ranking}
\centering
\small
\begin{tabular}{llrr}
  \hline
 Algorithm & Setup & \multicolumn{2}{c|}{Evaluations (Required Episodes) $\pm$ sd} \\ 
 &  & CartPole-v1 & MountainCar-v0 \\ 
  \hline
SMBNE & DynSet 5 ** &not tested& \it 130.67 $\pm$ 76.90 \\ 
SMBNE & DynSet 10 & \bf 57.57 $\pm$ 22.79 & 218.96 $\pm$ 129.70 \\ 
SMBNE & PreSet 5 & 63.17 $\pm$ 41.88 & 198.70 $\pm$ 152.10 \\ 
SMBNE & DynSet 5 & 64.83 $\pm$ 32.43 & \bf 179.40 $\pm$ 105.05  \\ 
SMBNE & InitSet 5 & 64.97 $\pm$ 34.43 & 327.22 $\pm$ 235.47  \\ 
SMBNE & DynSet 2 & 66.67 $\pm$ 43.98 & 320.67 $\pm$ 240.68  \\ 
SMBNE & LHS & 80.41  $\pm$  44.19 & 219.05 $\pm$ 152.43  \\ 

   \hline
CGP & MutRate 5 & \bf 328.00 $\pm$ 508.28 & 916.13 $\pm$ 1146.07 \\ 
CGP & MutRate 10 & 487.20 $\pm$ 626.35 & 1830.40 $\pm$ 1612.21 \\ 
CGP & MutRate 2 & 541.73 $\pm$ 897.50  & \bf 320.67 $\pm$ 240.68 \\ 
CGP & MutRate 1 & not tested  & 462.13 $\pm$ 667.98 \\ 
   \hline
RS & Base &\bf  1271.07 $\pm$ 1139.16 &\bf  5020.00 $\pm$ 0.00 \\ 
   \hline
\end{tabular}
\end{table}

Figure 3 and Table 2 show the results of all conducted experiments with both benchmark problems. For easier comparison, the results of the box plots are log10 scaled and colored according to the type of algorithm. The numbers indicate either the utilized CGP-ANN mutation rate in percent or the number of utilized state observation vectors $num_{s}$ for computing the PhD in SMB-NE.
\paragraph{CartPole-v1}
For CartPole-v1, all algorithms are able to find successful solutions, 
but show a high variation in solution quality over the different random seeds.
This variation relates to the different starting conditions for the RL environment and different initial data sets. 
On the one hand, the environment might rarely be solved by pure random chance during the initialization
of the algorithms. 
On the other hand, even CGP-ANN sometimes fails to find solutions within the specified budget of 3020 total fitness function evaluations.
In contrast, all SMB-NE variants are able to discover ANNs which solve these environments within the given budget.
The statistical tests indicate an overall significance of the results (Kruskal-Wallis rank sum test). 
However, no evidence for a significant differences between any CGP variant and Random Search is discovered by the statistical test procedure (posthoc), while all SMB-NE variants are evaluated to be different to CGP and Random Search. 
Between the tested inputs sets for SMB-NE, there is not sufficient evidence to indicate a significant difference according to the respective posthoc test. 
The results show that the best tested SMB-NE variants are able to clearly outperform the best tested CGP-ANN variant and require about 70\% (median) or 80\% (mean) fewer function evaluations (or environment episodes). 

\paragraph{MountainCar-v0}
As the results indicate, the MountainCar-v0 is more difficult to solve and CGP with random search is not able to discover a single valid solution. Overall, more evaluations are required to solve the task. 
The mutation rate in CGP-ANN has a noticeable influence on the performance.
The tested configuration with a mutation rate of 2\% performed best. 
Based on these result, additional runs with even smaller mutation rates were conducted (1\% is reported), but showed no improvements. 
For the sake of brevity, they are not shown in the result plots. 
Again, the Kruskal-Wallis rank sum test indicates that significant differences are present. 
The posthoc test shows that all algorithms performed statistical different from Random Search (with exception of CGP with 10\% mutation rate).
Only the SMB-NE \emph{DynSet**} variant, which features a more extensive surrogate model search, shows evidence for a significant difference to CGP-ANN. 
From the input sets, the \emph{Init} set variant performed worst. 
Still, the best standard dynamic variant \emph{DynSet 5} requires 45\% fewer evaluations than CGP-ANN and the dynamic variant \emph{DynSet**} performs even better (60-70\% less required function evaluations).

\paragraph{Discussion}
The presented experimental results provide substantial insights on the performance potential of utilizing model-based search in NE. For both test cases, SMB-NE outperforms the basic (1+4)-evolutionary strategy integrated in CGP-ANN.
First, we focus on the results of the model-free CGP-ANN with the (1+4)-ES. 
Particularly for MountainCar-v0, significant performance differences in the choice of mutation rate are visible.
This effectively shows the need for either exhaustive tuning of this parameter or development of an adaptive strategy.

Secondly, no clear statistical significant differences were observed for the different choices of how the state input vector for the PhD distance is generated. 
Thus, we are not able to support the different assumptions raised in the introduction of the input sets (Pre,Init, LHS and Dyn). 
Given the current experimental design with large input vectors, we can state that SMB-NE is fairly robust to the choice of the input vector. 
However, the MountainCar-v0 results show a slight preference towards the dynamic input sets, thus we still assume that it is preferable if the computed distance is based on state vectors that were actually observed, rather than artificially created or precomputed. 
Especially if a further dimension reduction of the PhD is considered, the dynamic input vector thus seems the best choice.

\begin{figure}[t]
    \centering
        \includegraphics[width=0.95\linewidth]{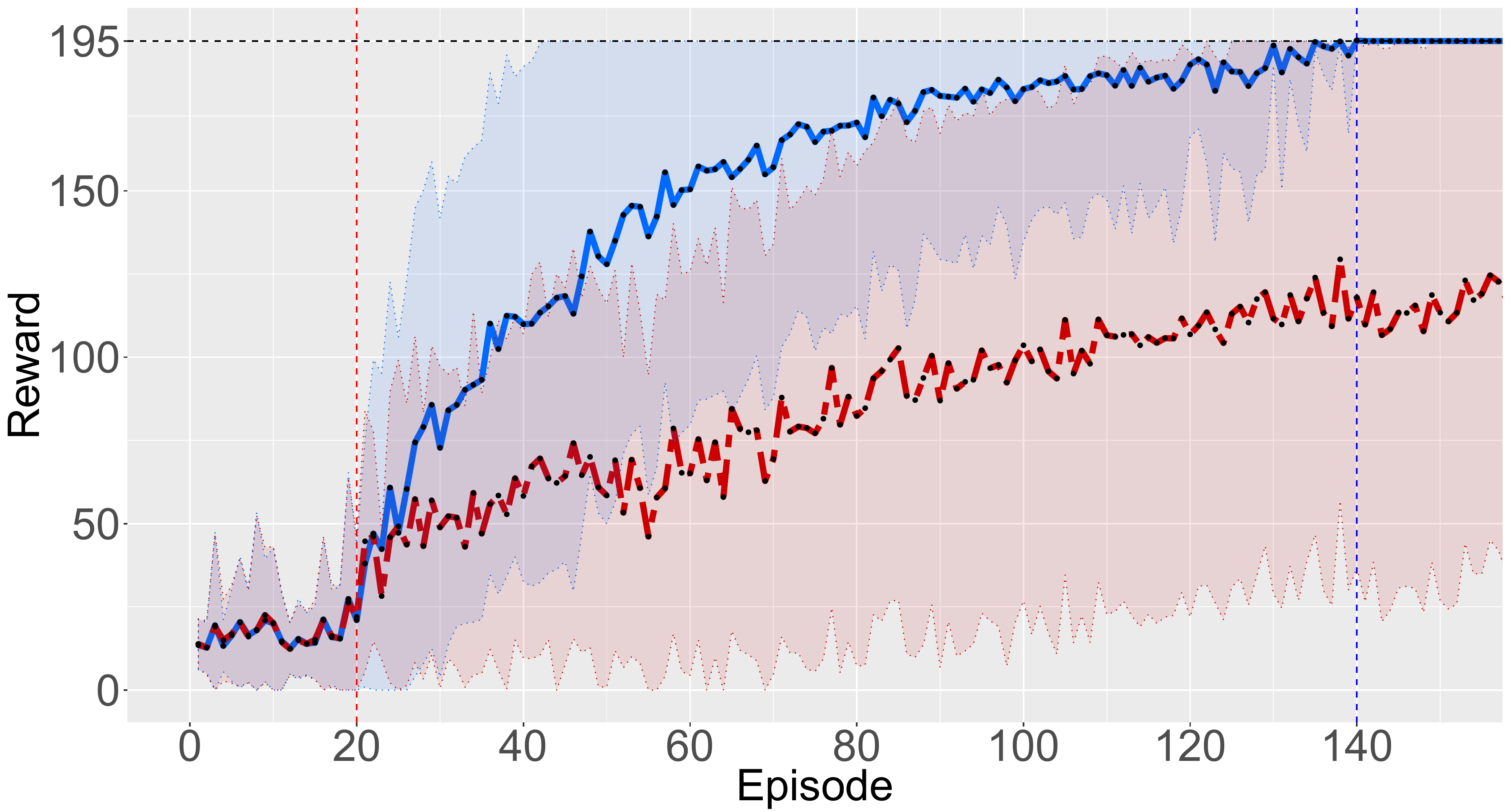}%
    \caption{Convergence plot of SMBNE.DynSet 10 (solid blue) and CGP.MutRate 5 (dashed red) on CartPole-v1, mean reward of current candidates (not best solution) aggregated over repeats with standard deviation (colored areas), the environment is solved by reaching a reward of 195 per episode.}
    \label{fig:convergence}
\end{figure}

In comparison to CGP-ANN, SMB-NE is in general able to produce more stable results, rendering it a promising choice for new tasks, as in the outlined scenario (S-1). 
For example, Figure \ref{fig:convergence} shows the convergence of the SMB-NE using a dynamic input vector with $num_{s}=5$ in comparison the CGP-ANN with a mutation rate of 5\%. 
The mean reward over the 30 repeats of the current candidate is shown for each iteration.
As can be observed, the model-based search shows a strong increase in reward after the initial set and then a steady convergence, while CGP-ANN improves also steady, but slower. 

\section{Conclusion and Future Work}\label{sec:conclusion}
In this work, we investigated how a surrogate model-based search can be utilized to enhance the efficiency of NE, given the complex task of evolving artificial neural networks for reinforcement learning. 
Our surrogate models are based on phenotypic distance measures which utilize the observed differences in the outputs of an ANN.
We discovered that our SMB-NE for RL is capable of significantly outperforming a model-free evolutionary strategy, which also answers our initially raised research question Q-II.
Regarding Q-I and Q-III, we proposed different approaches of generating state vectors for the ANN's input space. 
The current empirical results do not provide evidence for strong differences between the input sets generation methods, SMB-NE thus seems rather robust towards this choice. 
Still, we regard the dynamic input sets as the most promising approach.

Of course this work and the results raised further questions. 
The first is how to optimally set the parameters of the SMB-NE algorithm, 
particularly regarding the dimension of the input sets. 
Up to now, we do not know which length of state vector is required to generate a well-performing model and, thus, reasonable optimization performance. 
The length of the state vector is also related to the computation costs, particularly for computing the PhD measure.
The computation costs are a potential drawback of SMB-NE, but the clear and robust improvements of the evaluation efficiency render it notably attractive for tasks where the fitness evaluations themselves are very expensive. 
For instance, consider a robot controlled by an ANN. Testing that robot in a real environment may be very expensive, 
while computing only the outputs of the ANN are considerably cheaper.
In ongoing work we will furthermore attempt to generalize our results to more environments from the Gym toolkit as well as tests with real-world problems. 
We plan to investigate the underlying mechanics. 
In particular, the significance of certain algorithm parameters are of interest and may require more attention to algorithm tuning.

%

\bibliographystyle{ACM-Reference-Format}
\bibliography{Stor18d}

\end{document}